\documentclass{bmvc2k}
\usepackage{amsfonts}
\usepackage{multirow}
\usepackage{times}
\usepackage{epsfig}
\usepackage{graphicx}
\usepackage{amsmath}
\usepackage{amssymb}

\title{ST-ABN: Visual Explanation Taking into Account Spatio-temporal Information for Video Recognition}

\addauthor{Masahiro Mitsuhara}{mitsuhara@mprg.cs.chubu.ac.jp}{1}
\addauthor{Tsubasa Hirakawa}{hirakawa@mprg.cs.chubu.ac.jp}{2}
\addauthor{Takayoshi Yamashita}{takayoshi@isc.chubu.ac.jp}{2}
\addauthor{Hironobu Fujiyoshi}{fujiyoshi@isc.chubu.ac.jp}{2}

\addinstitution{
Department of Robotic Science and Technology\\
Graduate School of Engineering\\
Chubu University\\
1200 Matsumoto, Kasugai\\Aichi, 487-8501, Japan
}
\addinstitution{
Center for Mathematical Science and Artificial Intelligence\\
Chubu University\\
1200 Matsumoto, Kasugai\\Aichi, 487-8501, Japan
}

\runninghead{M. Mitsuhara, \etal}{Spatio-Temporal Attention Branch Network}

\def\etal{\emph{et al}\bmvaOneDot}

\begin{document}

\maketitle

\begin{abstract}
It is difficult for people to interpret the decision-making in the inference process of deep neural networks. Visual explanation is one method for interpreting the decision-making of deep learning. 
It analyzes the decision-making of 2D CNNs by visualizing an attention map that highlights discriminative regions. 
Visual explanation for interpreting the decision-making process in video recognition is more difficult because it is necessary to consider not only spatial but also temporal information, which is different from the case of still images. In this paper, we propose a visual explanation method called spatio-temporal attention branch network (ST-ABN) for video recognition. 
It enables visual explanation for both spatial and temporal information. 
ST-ABN acquires the importance of spatial and temporal information during network inference and applies it to recognition processing to improve recognition performance and visual explainability. 
Experimental results with Something-Something datasets V1 \& V2 demonstrated that ST-ABN enables visual explanation that takes into account spatial and temporal information simultaneously and improves recognition performance.
\end{abstract}

\section{Introduction}
Convolutional neural networks (CNNs)~\cite{LeCun1989,Alex2014}, which achieved higher image classification performance, have been applied to video recognition tasks. CNN-based video recognition methods can be categorized as either 2D CNN-based~\cite{simonyan2014two,feichtenhofer2016convolutional,feichtenhofer2017spatiotemporal,karpathy2014large,yue2015beyond,crasto2019mars,diba2017deep,lee2018motion,wang2016temporal,zhou2018temporal,lin2019tsm,girdhar2017actionvlad,li2020tea} or 3D CNN-based~\cite{hara2018can,tran2015learning,ji20123d,zolfaghari2018eco,xie2018rethinking,tran2018closer,qiu2017learning,carreira2017quo,feichtenhofer2019slowfast,zhao2018trajectory,wang2018videos,zhou2020spatiotemporal}. Most 2D CNN-based methods involve using a two-stream network structure~\cite{simonyan2014two,feichtenhofer2016convolutional,feichtenhofer2017spatiotemporal,yue2015beyond,girdhar2017actionvlad} where each video frame and an optical flow are input to two different CNNs. The two-stream methods then learns spatial and temporal information simultaneously by fusing spatial and temporal feature maps obtained from the two input networks. 3D CNN-based methods use a 3D convolution that extends the 2D convolution in the temporal direction. By stacking multiple 3D convolution layers~\cite{ji20123d,tran2015learning}, 3D CNNs can extract spatio-temporal features.

%
%
\begin{figure}[t]
\centering
\includegraphics[width=0.9\linewidth]{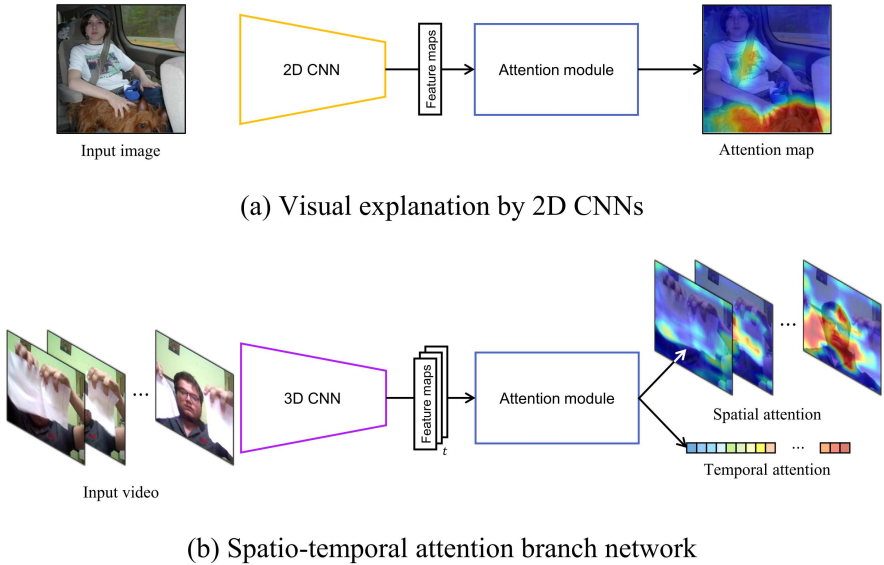}
\caption{Network structures of visual explanation by (a) 2D CNNs and (b) the proposed spatio-temporal attention branch network.}
\label{fig:topview_attention_map}
\end{figure}

These methods achieve high recognition performance. 
However, these methods suffer from understanding the decision-making process of the recognition results during inference, which is a common problem with CNN-based image recognition methods. Visual explanation methods have been widely investigated as image classification methods to interpret the basis of a CNN's decisions. 
As shown in Fig.~\ref{fig:topview_attention_map}(a), visual explanation methods~\cite{ribeiro2016should,chattopadhay2018grad,Selvaraju2017,smilkov2017smoothgrad,Zeiler2014,Zhou2016,fukui2018attention,montavon2018methods,Jost2015,fong2017interpretable,Petsiuk2018rise,fong2019understanding,jetley2018learn} analyze the decision-making of a 2D CNNs by visualizing an attention map that highlights discriminative regions. 
Typical visual explanation methods include class activation mapping (CAM) and gradient-weighted class activation mapping~(Grad-CAM). 
CAM outputs an attention map by using the response of the convolution layer. 
Grad-CAM outputs an attention map by using the positive gradients of a specific category. Although these methods enable us to interpret the basis for a CNN's decisions, such interpretation using CNN-based video recognition methods is still challenging. 
The reason is that video recognition requires not only spatial but also temporal information.

To solve this problem, we propose a visual explanation method for video recognition called \textit{spatio-temporal attention branch network}~(ST-ABN). Figure~\ref{fig:topview_attention_map}(b) illustrates an overview of the ST-ABN. 
It consists of 3D CNN and an attention module that outputs spatial and temporal attentions simultaneously. 
We introduce an  spatio-temporal~(ST) attention branch that takes into account both spatial and temporal information. 
Spatial attention highlights the importance of spatial information for each temporal segment. Temporal attention represents the importance of temporal information. We also apply an attention mechanism~\cite{Thang2015,Xu2015,Hu2017,Bahdanau2016,Volodymyr2014,Wang2017a,vaswani2017attention,Zichao2016,You2016,woo2018cbam,zhang2018attention,Wang2018b,chen2017sca} by weighting the acquired importance of spatial and temporal information to feature maps obtained from a feature extractor. 
This enables network training with a focus on significant spatio-temporal features for video recognition, improving recognition accuracy.

\section{Related Work}

%
In this section, we introduce CNN-based video recognition methods and visual explanations.

\subsection{Video Recognition}
CNN-based video recognition methods can be categorized as either 2D CNN-based~\cite{simonyan2014two,feichtenhofer2016convolutional,feichtenhofer2017spatiotemporal,karpathy2014large,yue2015beyond,crasto2019mars,diba2017deep,lee2018motion,wang2016temporal,zhou2018temporal,lin2019tsm,girdhar2017actionvlad,li2020tea} or 3D CNN-based~\cite{hara2018can,tran2015learning,ji20123d,zolfaghari2018eco,xie2018rethinking,tran2018closer,qiu2017learning,carreira2017quo,feichtenhofer2019slowfast,zhao2018trajectory,wang2018videos,zhou2020spatiotemporal}.

The 2D CNN-based methods apply 2D convolution to extract feature maps from each video frame and aggregate frame-by-frame information. 
These methods typically use a two-stream network structure~\cite{simonyan2014two,feichtenhofer2016convolutional,feichtenhofer2017spatiotemporal,yue2015beyond,girdhar2017actionvlad} in which each frame and the optical flow of motion information are input to two separate CNNs. 
Each network extracts spatial features from video frames and temporal features representing motion context from an optical flow. The spatial and temporal features are then fused. 
Temporal segment networks~(TSN)~\cite{wang2016temporal}, which are derived from these methods, learn in video units instead of frame units. 
A temporal relation network (TRN)~\cite{zhou2018temporal} replaces an average pooling operation that aggregates spatial and temporal, i.e., motion, features with an interpretable relational module. 
A temporal shift module (TSM)~\cite{lin2019tsm} can learn the temporal relationships between neighboring video frames without any additional computational costs. 
These 2D CNN-based methods fuse the spatial and temporal features extracted from different networks separately. 
Therefore, 2D CNN-based methods are vulnerable to discrepancy between two network outputs, and it is difficult to learn the inter-relationship between spatial and temporal information.

The 3D CNN-based video recognition methods use 3D convolution that extend a 2D convolution into spatial and temporal directions. 
The 3D CNN-based methods extract spatio-temporal features by stacking multiple 3D convolution layers~\cite{ji20123d,tran2015learning}. 
Unlike 2D CNN-based methods, the extracted spatio-temporal features take into account the inter-relationship between spatial and temporal information. 
Various 3D CNN-based methods have been proposed such as inflating 2D convolution kernels~\cite{carreira2017quo}, decomposing 3D convolution kernels~\cite{qiu2017learning,tran2018closer,xie2018rethinking,zhao2018trajectory}, and the slowfast networks~\cite{feichtenhofer2019slowfast} that is a two-stream network structure for extracting spatial features at a low temporal resolution and motion features at a high temporal resolution.

Video recognition methods that use an attention mechanism~\cite{Thang2015,Xu2015,Hu2017,Bahdanau2016,Volodymyr2014,Wang2017a,vaswani2017attention,Zichao2016,You2016,woo2018cbam,zhang2018attention,Wang2018b,chen2017sca} have also been proposed~\cite{Wang2018b,du2017rpan,chen20182,girdhar2017attentional,sharma2015action,xiao2019reasoning,zhu2016key}. 
Non-local neural networks~\cite{Wang2018b}, which are commonly used for introducing an attention mechanism, improve the accuracy of video recognition by capturing long-distance temporal dependency with a non-local operation capable of providing global information.

Although these 2D and 3D CNN-based video recognition methods have achieved high recognition performance, the decision-making process for network inference results is unclear. 
The difficulty in interpreting the video recognition decision-making process is that not only spatial but also temporal information needs to be taken into account.

\subsection{Visual Explanation}
Visual explanation~\cite{ribeiro2016should,chattopadhay2018grad,Selvaraju2017,smilkov2017smoothgrad,Zeiler2014,Zhou2016,fukui2018attention,montavon2018methods,Jost2015,fong2017interpretable,Petsiuk2018rise,fong2019understanding,jetley2018learn} is often used to interpret the decision-making of deep learning and has been widely investigated in image classification tasks. 
Visual explanation in an image classification task analyzes the decision-making of 2D CNNs by visualizing an attention map that highlights discriminative regions.

Visual explanation methods can be categorized into two approaches: gradient-based, which outputs an attention map using gradients, and response-based, which outputs an attention map using the response of the convolution layer. 
Grad-CAM~\cite{Selvaraju2017}, one of the gradient-based method, obtains an attention map for a specific category by using the response of the convolution layer and a positive gradient in the backpropagation process. 
Grad-CAM can be used for various pre-trained models. 
One of the response-based visual explanation methods is CAM~\cite{Zhou2016}, which outputs an attention map by using a $K$~channel feature map from the convolution layer of each category. The attention maps of each category are calculated using the $K$~channel feature map and the weight at a fully connected layer. 
However, CAM degrades recognition accuracy because spatial information is removed due to the global average pooling (GAP)~\cite{DBLP:journals/corr/LinCY13} layer between the convolution and fully connected layers. 
To solve this problem, the attention branch network (ABN) was proposed~\cite{fukui2018attention}, which extends an attention map for visual explanation to an attention mechanism. 
By applying an attention map to the attention mechanism, the ABN improves recognition performance and obtains an attention map simultaneously. 
These visual explanation methods can be applied to video to visualize spatial attention at each frame.

However, these attention maps are not referred to as the importance of temporal information. 
We aim to consider visual explanation of spatial and temporal information to video recognition.

\section{Proposed Method}

\begin{figure}[t]
\centering
\includegraphics[width=1.00\linewidth]{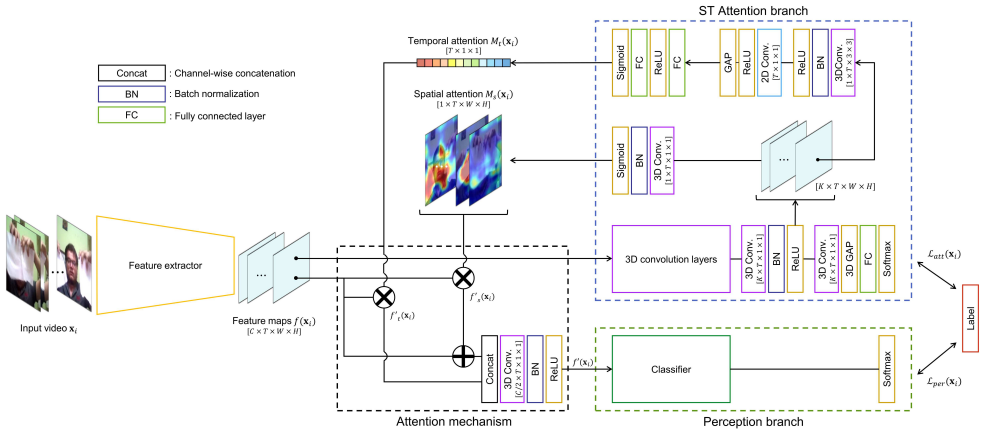}
\caption{Detailed structure of spatio-temporal attention branch network}
\label{fig:st-abn}
\end{figure}

In this paper, we propose a spatio-temporal attention branch network (ST-ABN), which enables visual explanation that takes into account the importance of spatial information and temporal information in video simultaneously. 
As shown in Fig.~\ref{fig:st-abn}, ST-ABN involves three modules: a feature extractor, spatio-temporal (ST) attention branch, and perception branch. 
The feature extractor consists of multiple convolution layers and outputs feature maps from the inputs. 
We introduce an ST attention branch that outputs spatial attention, which indicates the importance of spatial information, and temporal attention, which indicates the importance of temporal information, in a network based on 3D CNNs. 
The perception branch inputs feature maps with spatial and temporal attentions weighted on the feature maps by the attention mechanism and outputs the probability of each class.

\subsection{Spatio-Temporal Attention Branch}
As shown in Fig.~\ref{fig:st-abn}, the ST attention branch generates spatial attention that represents the importance of spatial information and temporal attention, indicating the importance of temporal information. 
The ST attention branch also outputs the classification results via 3D global average pooling (3D GAP). 
In the ST attention branch, the feature maps output from the feature extractor are first fed into the 3D convolution layers consisting of multiple residual blocks, which have the same structure as the perception branch. 
We set the stride of the convolution layer at the first residual block as 1 to maintain the resolution of the feature maps. 
The feature maps from the 3D convolution layers are then input to another 3D convolution layer of $K{\times}T{\times}1{\times}1$, where $K$ indicates the number of classes and $T$ indicates the number of frames. 
In other words, we can obtain $K$ feature maps for each frame. 
The feature maps, the size of which is $K{\times}T{\times}W{\times}H$, are input to the $K{\times}T{\times}1{\times}1$ 3D convolution layer, 3D GAP, and a softmax function. 
As a result, we can obtain the classification probabilities for each class.

\vspace{3.0mm}
\noindent
{\bf Spatial Attention}\hspace{3mm}
We generate spatial attention from the above-mentioned $K{\times}T{\times}W{\times}H$ feature maps. 
We apply a $1{\times}T{\times}1{\times}1$ 3D convolution layer for the $K{\times}T{\times}W{\times}H$ feature maps and obtain a single $1{\times}T{\times}W{\times}H$ feature map for each frame. 
This means that we aggregate the $K$ feature maps with respect to each video frame into a single feature map. 
After that, we can obtain a spatial attention map $M_s$ for each frame by applying a sigmoid function.

\vspace{3.0mm}
\noindent
{\bf Temporal Attention}\hspace{3mm}
Similar to spatial attention, we generate temporal attention from the $K{\times}T{\times}W{\times}H$ feature maps. 
These feature maps are first aggregated into a single $1{\times}T{\times}W{\times}H$ feature map with respect to each frame by applying a $1{\times}T{\times}3{\times}3$ 3D convolution layer. 
The channel dimensions of the $1{\times}T{\times}W{\times}H$ feature maps are then reduced and transformed into $T{\times}W{\times}H$ feature maps. 
The $T{\times}W{\times}H$ feature maps are further input to a $T{\times}1{\times}1$ 2D convolution layer, and $T{\times}W{\times}H$ feature maps are obtained. 
We calculate the mean value of each feature map in the spatial direction by the convolution layer with the number of channels for the number of frames and GAP. 
Finally, the temporal attention is generated via the fully connected layer, rectified linear unit (ReLU), and sigmoid functions. 
Here, we adopt a simple gating mechanism that uses sigmoid functions such as squeeze-and-excitation networks (SENet)~\cite{Hu2017}. 
This enables our network model to emphasize multiple frames instead of only a single frame.

\subsection{Attention Mechanism}
The spatial and temporal attentions acquired from the ST attention branch is further used as an attention mechanism for weighting feature maps. 
Let ${\bf x}_i$ be the $i$-th sample in a dataset and $f\left( {\bf x}_i \right)$ be the corresponding feature maps obtained from the feature extractor. 
The weighted feature maps $f_s'({\bf x}_i)$ by spatial attention $M_s({\bf x}_i)$ is defined as
\begin{equation}
f_s'({\bf x}_i)=\left( 1+M_s \left({\bf x}_i\right) \cdot f\left( {\bf x}_i \right) \right).
\end{equation}
For spatial attention, we apply a residual mechanism~\cite{Wang2017a} and add the unweighted feature maps to the weighted feature maps. 
This can suppress the disappearance of the feature maps, and the attention maps can be efficiently reflected in the recognition. 
The attention mechanism with the temporal attention $M_t({\bf x}_i)$ calculates the weighted feature maps $f_t'({\bf x}_i)$ as
\begin{equation}
f_t'({\bf x}_i)=M_t\left({\bf x}_i\right) \cdot f\left( {\bf x}_i \right).
\end{equation}
For the temporal attention mechanism, we apply simple weighting and do not use a residual attention mechanism.

These $f_s'({\bf x}_i)$ and $f_t'({\bf x}_i)$ are combined in the channel direction by
\begin{equation}
f'({\bf x}_i)=\mathrm{conv}_{\theta}\left( {\rm concat}[f'_s({\bf x}_i) , f'_t({\bf x}_i)] \right)
\end{equation}
where $f'({\bf x}_i)$ denotes the concatenated feature maps. 
The number of channels is doubled because the two feature maps are channel-wise concatenated. The concatenated $f'({\bf x}_i)$ are then integrated by applying the 3D convolution layer of $C/2{\times}T{\times}1{\times}1$ that halves the number of channels of the concatenated feature maps. 
These integrated feature maps are inputted into the perception branch, and the final recognition results are obtained. This enables learning focusing on important spatio-temporal features.

\subsection{Training}
The loss function of ST-ABN $\mathcal{L}({\bf x}_i)$ is calculated as
\begin{equation}
\mathcal{L}({\bf x}_i)=\mathcal{L}_{att}({\bf x}_i)+\mathcal{L}_{per}({\bf x}_i),
\end{equation}
where ~$\mathcal{L}_{att}({\bf x}_i)$ denotes the training loss at the ST attention branch, and ~$\mathcal{L}_{per}({\bf x}_i)$ denotes the training loss at the perception branch. 
The $\mathcal{L}_{att}({\bf x}_i)$ and $\mathcal{L}_{per}({\bf x}_i)$ can be calculated by the softmax function and cross-entropy error. 
The loss function of ST-ABN is trained in an end-to-end manner.

\subsection{Implementation Details}
In this section, we explain the implementation of ST-ABN. ST-ABN is constructed by dividing a backbone network into a feature extractor and perception branch and adding an ST attention branch between the feature extractor and perception branch. 
Therefore, it can be easily introduced into various network models (e.g., C3D, 3D ResNet). 
In this study, we used 3D ResNet, which is a temporal direction inflated version of ResNet~\cite{He2016}, as the backbone network. 
Specifically, ST-ABN is constructed using 3D ResNet based on the slow pathway of slowfast networks~\cite{feichtenhofer2019slowfast}. 
The spatial dimension of the input is 224$\times$224, and input data size is $C{\times}T{\times}W{\times}H$. 
We apply a dropout~\cite{hinton2012improving} of 0.5 for the ST attention branch and perception branch to suppress overfitting.

\begin{table}[t]
\centering
{\footnotesize \tabcolsep = 3.7mm
\scalebox{0.8}{
\begin{tabular}{llccccc}
 & & & \multicolumn{2}{c}{Something. V1} & \multicolumn{2}{c}{Something. V2} \\ \cline{4-7}
Method & Backbone & Frames & Top-1 & Top-5 & Top-1 & Top-5 \\ \hline \hline
TSN~\cite{wang2016temporal} & BN-Inception & 8 & 19.5 & -- & 27.8 & 57.6 \\
TRN Multiscale~\cite{zhou2018temporal} & BN-Inception & 8 & 34.4 & -- & 48.8 & 77.6 \\
MFNet-C101~\cite{lee2018motion} & ResNet-101 & 10 & 43.9 & 73.1 & -- & -- \\
CPNet~\cite{liu2019learning} & ResNet-34 & 24 & -- & -- & 57.7 & 84.0 \\
TSM~\cite{lin2019tsm} & ResNet-50 & 8 & 45.6 & 74.2 & 59.1 & 85.6 \\
TSM~\cite{lin2019tsm} & ResNet-50 & 16 & 47.2 & 77.1 & 63.4 & 88.5 \\
TSM$_{En}~$\cite{lin2019tsm} & ResNet-50 & 24 & 49.7 & 78.5 & -- & -- \\
STM~\cite{jiang2019stm} & ResNet-50 & 8 & 49.2 & 79.3 & 62.3 & 88.8 \\
STM~\cite{jiang2019stm} & ResNet-50 & 16 & 50.7 & 80.4 & 64.2 & 89.8 \\
GST~\cite{luo2019grouped} & ResNet-50 & 16 & 48.6 & 77.9 & 62.6 & 87.9 \\
ABM~\cite{zhu2019approximated} & ResNet-50 & 16$\times$3 & 46.8 & -- & 61.3 & -- \\
DFB-Net~\cite{martinez2019action} & ResNet-152 & 16 & 53.4 & 81.8 & 57.7 & 84.0 \\
bLVNet-TAM~\cite{NEURIPS2019_3d779cae} & bLResNet-101 & 32$\times$2 & 53.1 & 82.9 & 65.2 & 90.3 \\
TEA~\cite{li2020tea} & ResNet-50 & 16$\times$3$\times$10 & 52.3 & 81.9 & -- & -- \\
SmallBig$_{En}$~\cite{li2020smallbignet} & ResNet-50 & 24$\times$2$\times$3 & 51.4 & 80.7 & 64.5 & 89.1 \\
PEM~\cite{weng2020temporal} & ResNet-50 & 16$\times$2 & 52.0 & -- & 65.0 & -- \\ \hline

I3D~\cite{wang2018videos} & 3D ResNet-50 & 64 & 41.6 & 72.2 & -- & -- \\
I3D~\cite{xie2018rethinking} & 3D BN-Inception & 64 & 45.8 & 76.5 & -- & -- \\
Non-local I3D + GCN~\cite{wang2018videos} & 3D ResNet-50 & 64 & 46.1 & 76.8 & -- \\
ECO~\cite{zolfaghari2018eco} & 3D BN-Inception + 3D ResNet-18 & 8 & 39.6 & -- & -- & -- \\
ECO$_{En} Lite$~\cite{zolfaghari2018eco} & 3D BN-Inception + 3D ResNet-18 & 92 & 46.4 & -- & -- & -- \\
S3D-G~\cite{xie2018rethinking} & Inception & 64 & 48.2 & 78.7 & -- & -- \\
TrajectoryNet~\cite{zhao2018trajectory} & 3D ResNet-18 & -- & 47.8 & -- & -- & -- \\
Zhou {\it et al.}~\cite{zhou2020spatiotemporal} & 3D DenseNet-121 & 16 & 50.2 & 78.9 & 62.9 & 88.0 \\ \hline

ST-ABN & 3D ResNet-50 & 32 & 45.9 & 75.4 & 56.6 & 82.0 \\
ST-ABN & 3D ResNet-50 & 32$\times$2 & 53.3 & 82.0 & 64.1 & 89.6 \\
ST-ABN & 3D ResNet-101 & 32 & 47.4 & 76.4 & 58.0 & 83.2 \\
ST-ABN & 3D ResNet-101 & 32$\times$2 & {\bf55.0} & {\bf83.0} & {\bf65.8} & {\bf90.4} \\
\end{tabular}
}
}
\vspace{1mm}
\caption{Performance evaluation~(top-1 and top-5 accuracies) of conventional methods and ST-ABN for Something-Something datasets V1 \& V2 [\%]}
\label{tab:acc}
\end{table}
\section{Experiments}
We evaluated the effectiveness of ST-ABN using Something-Something datasets V1 \& V2, which are benchmarks for action recognition. 
We first compared the recognition accuracy of ST-ABN, with those of conventional methods. 
We also qualitatively and quantitatively evaluated the explainability of spatial and temporal attentions.

\subsection{Datasets}
Something-Something datasets V1 \& V2~\cite{goyal2017something} are used as benchmarks for large-scale action recognition, and they recognize 174 basic actions of a person handling everyday objects. 
The length of the videos ranges from 2 to 6 seconds. 
Something-Something dataset V1 contains 108,499 videos, and the training, validation, and evaluation data contain 168,913, 24,777, and 27,157 videos, respectively. 
Something-Something dataset V2 is a dataset that expands the number of videos in V1 by more than 2 times and contains 220,847 videos. 
The training, validation, and evaluation data contain 168,913, 24,777, and 27,157 videos, respectively. 
We conducted experiments using the training and validation data from these datasets.

\subsection{Experiment Details}
We used Something-Something datasets V1 \& V2 to compare the recognition accuracies among conventional methods and ST-ABN. The backbone networks of ST-ABN were 3D ResNet-50 and 3D ResNet-101. As for the number of frames to be input, the recognition accuracies were compared between the case of inputting 32 frames selected at random and that of inputting two 32 frames in which the input video is randomly divided into two segments. We optimized the networks by stochastic gradient descent (SGD) with momentum and set a momentum and weight decay of 0.9 and 0.0005, respectively. We used over 8 GPUs, and each GPU had a batch-size of 8, resulting in a mini-batch of 64 in total. Our models were initialized using pre-trained models on ImageNet~\cite{Deng2009}. All models started training at a learning rate of 0.01, and the learning rate was multiplied by $1/10$ after the saturation of the validation loss.

\begin{figure}[t]
\centering
\includegraphics[width=1.00\linewidth]{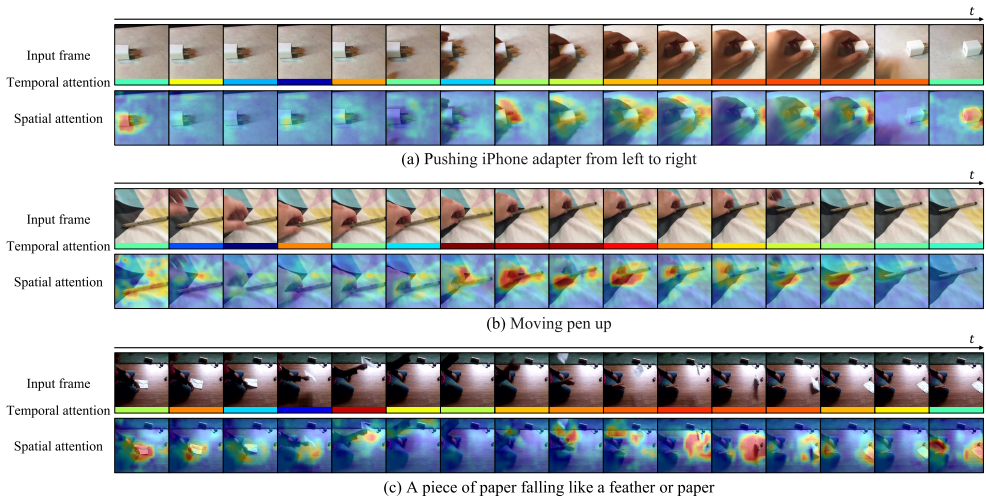}
\caption{Visualization results of spatial attention and temporal attention for Something-Something dataset V2. Figures (a), (b), and (c) show the obtained attentions for video samples. From top to bottom, each figure shows input video frames, the corresponding temporal attention, spatial attention, and action class.}
\label{fig:att-map}
\end{figure}

\subsection{Comparison with Conventional Methods}
\label{sec:comparison}
We compared the performances of various conventional methods and ST-ABN. 
ST-ABN performed better than the state-of-the-art methods. 
Table~\ref{tab:acc} shows the top-1 and top-5 accuracies of a comparison between the conventional methods and ST-ABN for Something-Something datasets V1 \& V2. 
ST-ABN performed the best, indicating that recognition accuracy can be improved by applying spatial attention and temporal attention to the recognition process through the attention mechanism.

\subsection{Evaluation of Attention}
We qualitatively and quantitatively evaluated the visual explanation of the spatial and temporal attentions.

\vspace{3.0mm}
\noindent
{\bf Qualitative Evaluation}\hspace{3mm}
We visualized spatial and temporal attentions through qualitative evaluation. Figure~\ref{fig:att-map} shows examples of visualized spatial and temporal attentions. 
For spatial attention, we visualized the attention maps of each frame as heat maps. 
The temporal attention of each frame was visualized as the colors of a heat map. 
The color bar corresponding to each frame is a color representation of the weight of the temporal attention.

From the visualization results of spatial attention, spatial attention strongly highlights the hand regions handling any object and weakly highlights the object to be handled.

From the results of temporal attention, large weight outputs correspond to frames with the motion representing the class of action recognition. 
Furthermore, frames in which action took place were continuously outputting large weight values. 
This temporal attention provides intuitive insights into the attention mechanism of ST-ABN.

These results indicate that ST-ABN can enable visual explanation that takes into account both spatial and temporal information simultaneously.

\begin{table}[t]
\centering
{\footnotesize \tabcolsep = 6.0mm
\scalebox{1.0}{
\begin{tabular}{cccccc}

\multicolumn{2}{c}{Attention} & \multicolumn{2}{c}{Something-Something V1} & \multicolumn{2}{c}{Something-Something V2}  \\ \hline
Spatial & Temporal & Top-1 & Top-5 & Top-1 & Top-5 \\ \hline \hline
 &  & 53.3 & 82.0 & 64.1 & 89.6 \\
\checkmark &  & 28.8 & 57.8 & 32.3 & 60.2 \\
 & \checkmark & 20.0 & 50.6 & 43.1 & 72.5 \\
\checkmark & \checkmark & {\bf 6.9} & {\bf 24.3} & {\bf 14.2} & {\bf 33.5} \\ 
\end{tabular}
}
\vspace{1mm}
\caption{Comparison of top-1 and top-5 accuracies when spatial and temporal attentions were inverted for Something-Something datasets V1 \& V2 [\%]}
\label{tab:inverse}
}
\end{table}

\vspace{3.0mm}
\noindent
{\bf Quantitative Evaluation}\hspace{3mm}
To quantitatively evaluate the explanatory nature of spatial and temporal attentions, these attentions were reversed. 
We then carried out inference with the inverted spatial and temporal attentions were then carried out and compared the recognition accuracy of ST-ABN were compared with and without inverted attention to confirm the effectiveness of spatial and temporal information for recognition. 
Spatial attention and temporal attention are inverted by
\begin{equation}
M_{\mathrm{inverse}}({\bf x}_i)=1-M({\bf x}_i),
\end{equation}
where $M({\bf x}_i)$ represents spatial and temporal attentions, and $M_{\mathrm{inverse}}({\bf x}_i)$ represents inverted spatial and temporal attentions. 
In this experiment, we compared four conditions: no reversal, reversal of spatial attention only, reversal of temporal attention only, and reversal of spatial and temporal attentions.

Table~\ref{tab:inverse} shows the results of a comparing the recognition accuracy of ST-ABN by reversal of spatial and temporal attentions for Something-Something datasets V1 \& V2. 
The recognition accuracy decreased when only spatial attention was reversed and when only temporal attention was reversed. 
Reversing both spatial and temporal attentions drastically reduced recognition accuracy. Therefore, effective attention regions for recognition were obtained because the recognition accuracy substantially decreased by reversing the spatial and temporal attentions.

\section{Conclusion}
In this paper, we proposed a spatio-temporal attention branch network (ST-ABN) for video recognition for visual explanation for both spatial and temporal information.
ST-ABN acquires the importance of spatial and temporal information during the inference of 3D CNN-based models, which can be applied to the attention mechanism to improve the visual explanation and recognition performance.
Experimental results with Something-Something datasets~V1 \& V2 showed that ST-ABN improves the top-1 and top-5 accuracy compared with conventional methods.
In a qualitative evaluation of spatial and temporal attentions, ST-ABN enabled visual explanation that takes spatial and temporal information into account simultaneously.
In a quantitative evaluation of spatial and temporal attentions, we demonstrated that reversing spatial and temporal attentions significantly reduced the recognition accuracy and obtained effective attention regions for recognition.
Our future work is to extend ST-ABN for other video recognition tasks.

\bibliography{refs}
\end{document}